\documentclass[english]{IEEEtran}
\usepackage[T1]{fontenc}
\usepackage[latin9]{inputenc}
\usepackage{float}
\usepackage{graphicx}

\makeatletter

\providecommand{\tabularnewline}{\\}

\usepackage{url}

\makeatother

\usepackage{babel}
\begin{document}

\title{NIMBUS: A Hybrid Cloud-Crowd Realtime Architecture for Visual Learning
in Interactive Domains}

\author{Nick DePalma and Cynthia Breazeal\\
Personal Robots Group\\
MIT Media Lab\\
20 Ames Str.\\
Cambridge, MA 02139}
\maketitle
\begin{abstract}
Robotic architectures that incorporate cloud-based resources are just
now gaining popularity \cite{thielman2015}. However, researchers
have very few investigations into their capabilities to support claims
of their feasibility. We propose a novel method to exchange quality
for speed of response. Further, we back this assertion with empirical
findings from experiments performed with Amazon Mechanical Turk and
find that our method improves quality in exchange for response time
in our cognitive architecture.
\end{abstract}

\section{Introduction}

Real, working, robotic systems must incorporate systemic architecture
designs. The simple heterogeneity present in autonomous electro-mechanical
(and biological) systems means that interfaces between differing subsystems
are inevitable consequences of assembly. But while cognitive architecture
research has, by definition, been inspired by biology, neuroscience,
cognitive science, or psychology, we must also realize that robots
are connected to resources that humans and animals don't have access
to. Wireless and network connectivity allows many robots lives to
be connected to internet resources to support the interaction. 

Our software framework has been built over the past few decades around
the idea of synthetic lifelike agents \cite{downie2001creature},
and was adapted into a more embodied approach that takes into account
sensors and motors and incorporates elements of interactive task learning
\cite{thomaz2006socially} and principles of animation \cite{breazeal2003interactive}.
In our recent past, our architecture has forked into two separate
initiatives: one in which emphasizes the ability to take advantage
of small, embedded, and ever improving smartphones \cite{williams2013reducing,williams2013towards,gordon2015can}
and an architecture that emphasizes its ability to operate across
small and large scale computer systems \cite{depalma2011leveraging,breazeal2013crowdsourcing}.
This paper will focus on a simple pipeline to unify these movements
to support lightweight calls to resources not available to the robot
alone. With this innovation, the robot may make decisions to connect
with individuals on the internet based on the availability of resources
and participants not present for the agent interacting with the world.

Our server, Nimbus%
\footnote{This resource is always online at http://nimbus.media.mit.edu%
}, operates as an always available resource for robots in the field
and for interaction partners. Nimbus serves to collect and coordinate
information from an always-on and resilient data collection interface,
crowdsource on-demand when absolutely needed by the robot, and to
act as a computational resource for highly intensive processes. This
paper documents underlying questions regarding speed of response and
security that are highly relevant to interactive domains like human-robot
interaction. 

The trade-off between quality of response from people on the Internet
and the speed at which a response can be obtained by an autonomous
system are still not well understood. Our research group over the
past few years has built a resilient cloud infrastructure to support
the robots on the ground and in the field. One of the higher level
features of the system is to retrieve information from databases in
the cloud and when unavailable, obtain high quality labels from crowds
of people on systems such as Amazon Mechanical Turk or Crowd Flower.
In this paper, we measure round trip times from the robot to the crowd
and back and discuss what one can expect from a system that leverages
crowdsourcing as a core cognitive resource. We introduce a novel mechanism
to improve quality of the response at the expense of response speed.
We conclude with remarks and thoughts about the future of technology
built for the purpose of efficient and interactive labeling to support
interaction.

\subsection*{Related Work}

\begin{figure*}[t]
\begin{centering}
{\huge{}\includegraphics[width=0.72\paperwidth]{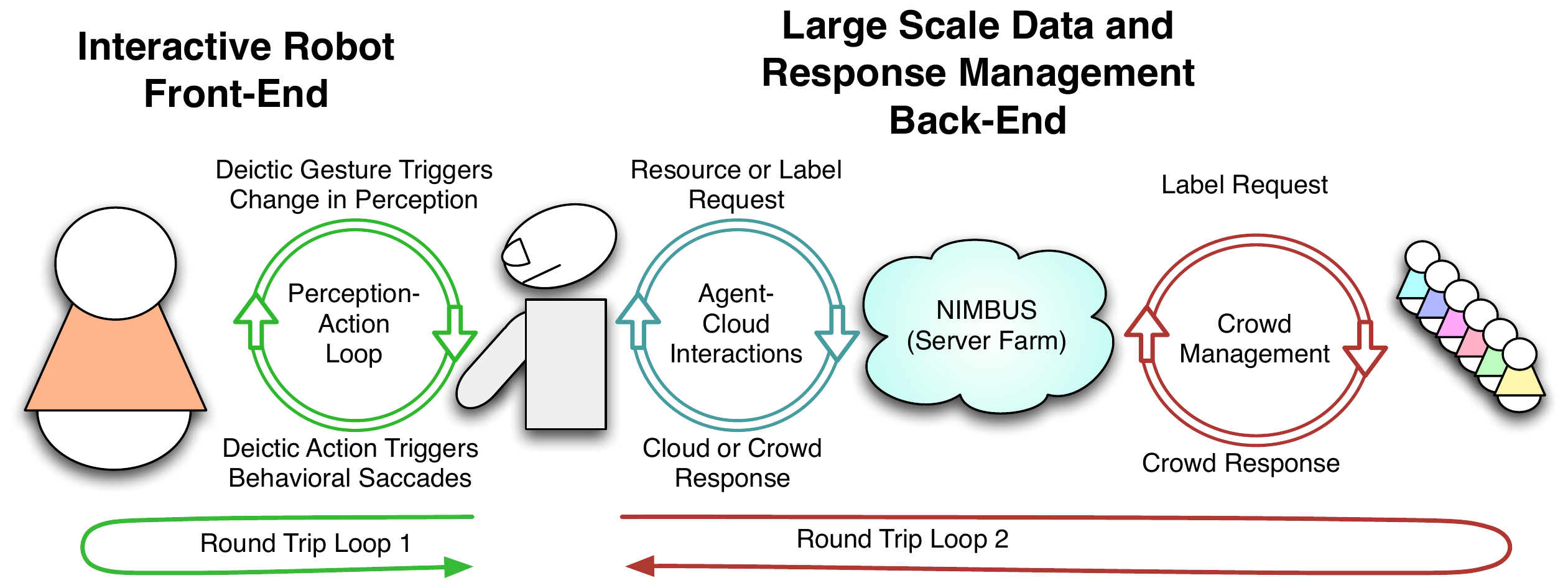}}
\par\end{centering}{\huge \par}

\centering{}\caption{Cognitive Architecture for Visual Learning Problem}
\label{fig:cogarch}
\end{figure*}

Few architectures have integrated real-time crowdsourcing and cloud-based
resources into their cognitive architectures and if these architectures
have, they have not been integrated into socially intelligent robotics
tasks to support these dynamic interactions. Many surveys cover the
broad spectrum of research that utilizes crowdsourcing in robotic
learning tasks that range from perception learning \cite{russell2008labelme}
to action learning \cite{sorokin2010people}. For a survey on various
projects that use crowdsourcing in machine learning and robotic tasks,
see the comprehensive reviews by Kehoe et. al. \cite{goldberg2013cloud,kehoe2015survey}.
For example, the RoboEarth project vision is still the most mature
vision of cloud resources integrating backend architectures to support
perception, action, and mapping \cite{roboearth2016}. Rapyuta, the
underlying framework, can be very powerful but it's emphasis on non-interactive
tasks makes real-time latency issues secondary for their level of
inquiry \cite{hunziker2013rapyuta}. Details on many of these architectures
are sparse, with Rapyuta \cite{hunziker2013rapyuta} and the work
by Osentoski et. al. utilizing closed-loop control \cite{osentoski2010crowdsourcing}
with a single online participant providing the most concrete details.
Other architectures treat crowdsourcing as a secondary or tertiary
characteristic of the research process, performed after data collection
to support annotation. 

The research previously mentioned (except perhaps \cite{hunziker2013rapyuta})
does not utilize the cloud as a resource it can autonomously decide
to use when other resources are unavailable. Rapyuta manages to schedule
available resources by the cloud and the real-time system using specialized
scheduling algorithms in which large tasks can be scheduled to operate
on cloud based hardware or on local embedded systems inside of the
robot depending on the needs and demands of the entire architecture.
Our system, instead, focuses on the labeling task as a source of information
that can be made available on-demand through either a social interaction
between the interaction partner and the robot or as an interaction
it can request online. 

In this work we use this crowdsourcing system to support interactive
perceptual learning between the robot and its environment. To support
this level of inquiry, we also review real-time perceptual learning
systems. Raptor \cite{goehring2014interactive} is a recent (and closely
related) successful dynamic training process that utilizes the power
of ImageNet\cite{deng2009imagenet}, a large database of object categories,
to discover early image priors that it can use as in-situ training
instances. ImageNet is used entirely for figure-ground patch discovery.
These image patches can then be used to train fast whitened HOG-based
object detectors dynamically throughout the interaction. This process
takes under 30 seconds for new image patch to reify into an object
detector. Our perception-action loop extends this type of system and
allows the robot to take deictic actions toward exchanging figure-ground
hypothesis enumeration with a social partner to support object discovery.
This paper serves to support this perception-action loop (documented
more thoroughly in \cite{depalma2015discovery,depalma2015sensorimotor,depalma2016bootstrapping})
and focuses primarily on incorporating crowd based labels into this
cognitive architecture.

\section{On-the-fly Visual Labeling from Local and Remote Sources}

Our system seeks to realize a simple abstraction: that perception-action
loops help define our visual labeling task interactively and that
backend loops completed between the real-time embedded system and
our servers are present to support the needs of the interactive system
(see Figure \ref{fig:cogarch}). We have chosen a visual labeling
task as this is an area where early stimuli need to be reified into
abstract representations that can be reused within the interaction.
Interactive visual labeling in which a system is able to take early
visual stimuli and dynamically create stable object detectors (or
more abstractly deictic pointers\cite{brooks2006working,ballard1997deictic})
in real time is still a large open problem space that is rich with
potential impact.

To demonstrate the power of this system, we have built a system to
learn labels both through an interaction and through our server architecture
that manages an Amazon Mechanical Turk account. Figure \ref{fig:cogarch}
shows the cognitive architecture that surrounds our system. At the
interaction level, the system utilizes gaze and deictic gesture to
visually guide the robot towards relevant areas in the scene. It uses
finite horizon reasoning and simple low-level cues to visually extract
select image patch foregrounds from the scene. This interactive guiding
process allows the robot to be directed towards stimuli it wouldn't
normally be directed towards if it operated in a purely discretized
object-oriented manner and to additionally use those early stimuli
as training instances for socially relevant deictic and lexical referencing.
In the event that the user walks away, the robot still has the ability
to saccade to new stimuli and collect labels from Amazon Mechanical
Turk for a small fee.

The following two sections introduce a natural tension in the crowdsourcing
architecture between the trade off of speed vs quality. We introduce
two methods to document the trade off that we can expect out of systems
of this type. One method is what we call one-shot labeling. In one-shot
labeling the system attempts to collect the label as fast as possible.
The other method, called Rollover Labeling, attempts to reliably get
a label in the presence of noisy, error prone, crowd responses by
rolling over the result from one crowdsourcing participant to another
in an attempt to correct and clarify previous labels. 

\begin{figure*}
\centering{}{\huge{}\includegraphics[width=0.75\paperwidth]{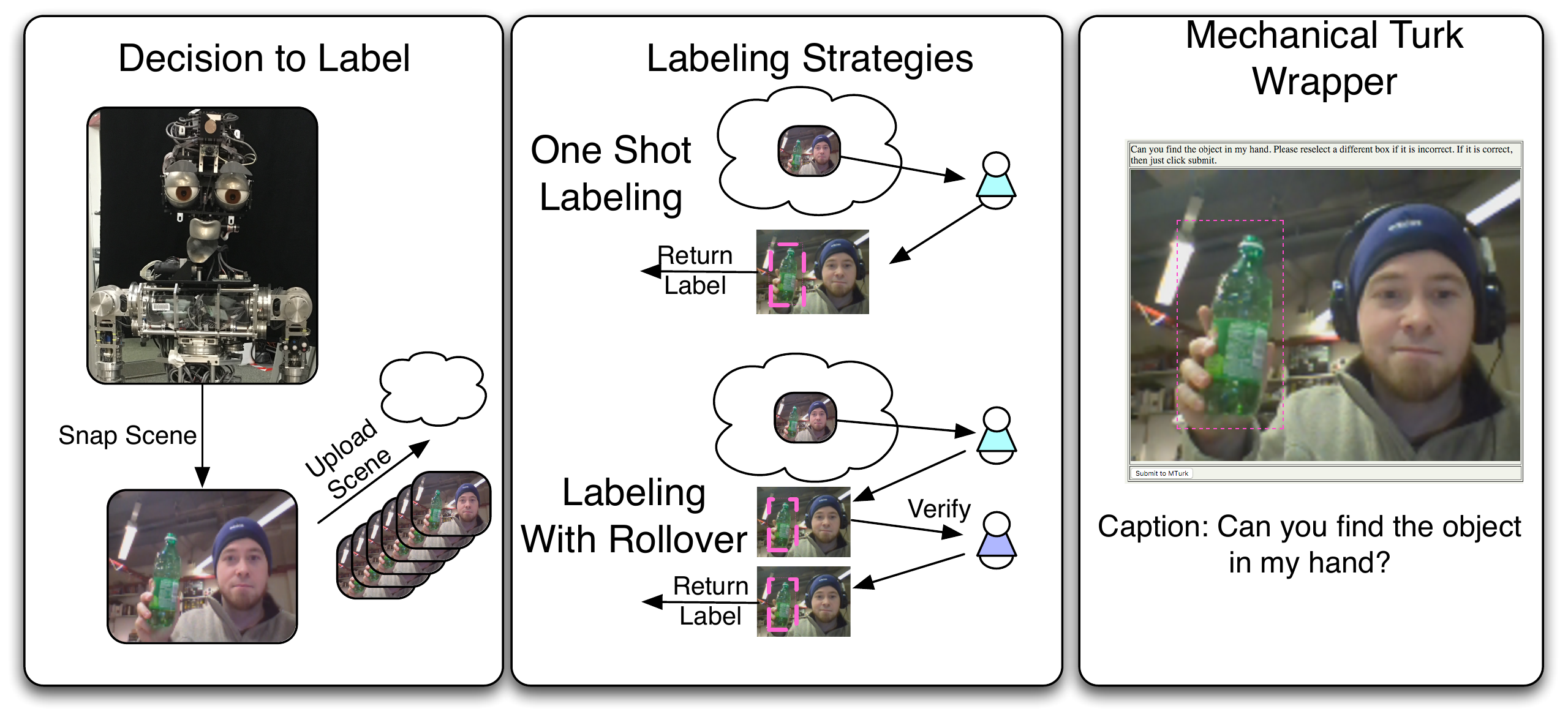}}
\caption{Data Collection Procedure}
\label{fig:collection}
\end{figure*}

\section{Time-to-Response Measurements}

\begin{figure}[h]
\includegraphics[width=1\columnwidth]{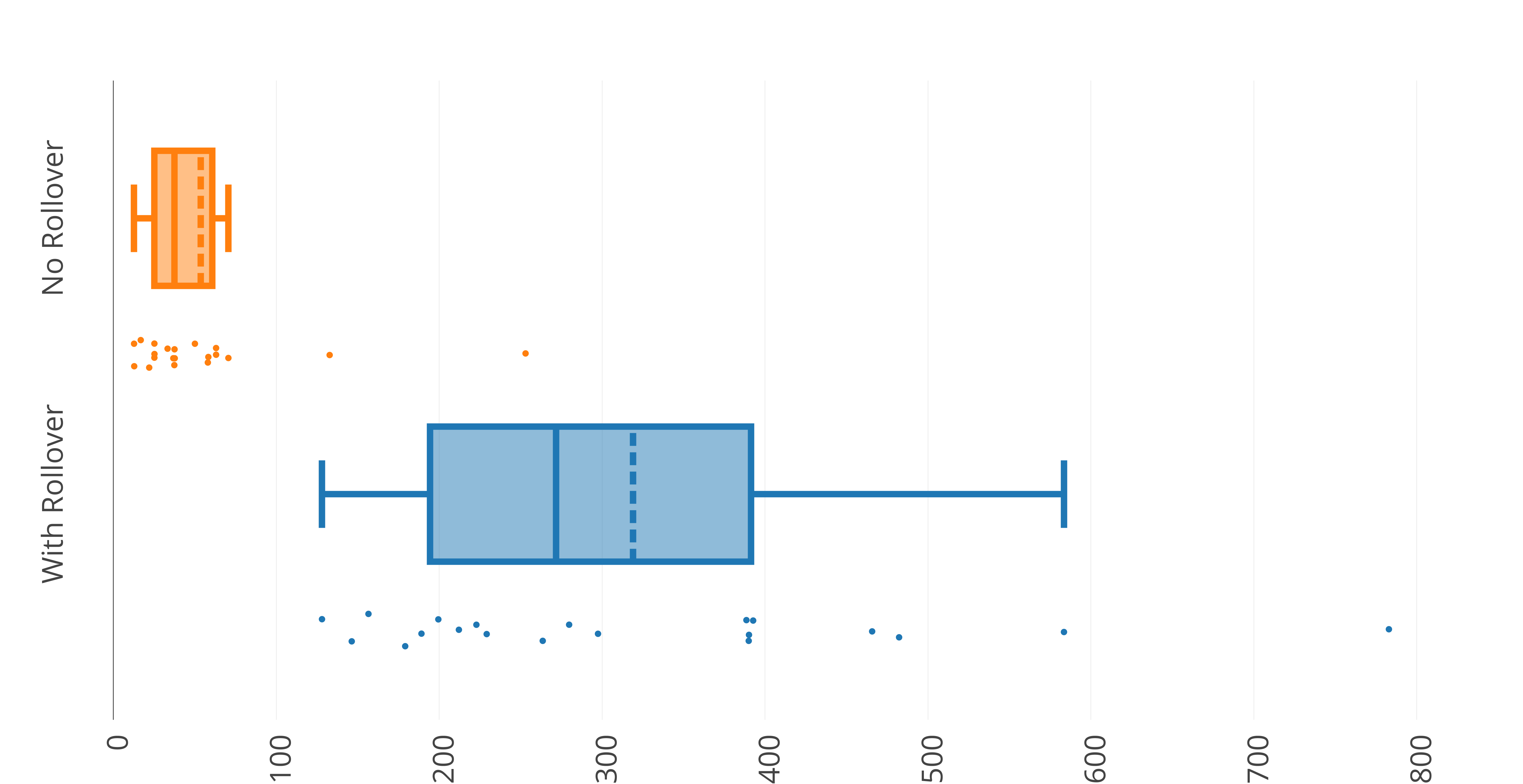}

\caption{Response times of two backend data collection strategies in seconds.}
\label{fig:timetoresponse}
\end{figure}

What's the shortest latency that one could expect if the robot needs
the label as quickly as possible? And what do we lose if we want the
highest quality response? In our case, our experience is that to use
crowdsourcing as an on-demand resource, the robot must:
\begin{enumerate}
\item take a snapshot from its sensors
\item upload the data and wrap it in a visually legible and descriptive
manner to a server where it can be reliably served to a participant
online
\item find a participant to label it for a price%
\footnote{This seems to be the slowest process.%
}
\item wait for the participant to accept the job and read the instructions
\item await the response
\end{enumerate}
We used our architecture to snap a picture to label from a webcam
on a simple robot (see Figure 2). The image and requisite interface
details were uploaded to Nimbus and a process began to synchronously
wait for a response. Total round-trip times are reported in Figure
\ref{fig:timetoresponse}. Our process used Amazon Mechanical Turk
at a competitive rate of \$0.25 for a 45 second labeling task. We
ran the experiment 20 times for each strategy.

Figure \ref{fig:timetoresponse} shows that one-shot labeling can
be a relatively quick process. Response times for one-shot labeling
fell in the range of $[12.9s, 520s]$ with a median of $37.5s$ while
a rollover process fell in the range of $[95s, 1217.8s]$ with a median
of $263.58$ seconds, or approximately $4.3$ minutes. It is clear
that one-shot labeling has the potential to be a very fast process,
optimistically only taking about $13$ seconds!

\section{Quality-Speed Tradeoff}

On the other hand, what kind of data quality can you expect out of
quick 30 second one-shot label acquisition? In previous experiments\cite{depalma2011leveraging},
we found that the quality of online data can vary widely. Using the
same results that we acquired from our 20 participant online experiment,
we calculated the mean bounding box and use it as a baseline answer.
We calculated the mean squared error (MSE) with respect to the baseline
and found that rollover significantly reduced the overall error. Additionally,
we saw that the responses from rollover provided a reduction in response
variance. This suggests that the participants' extra time spent on
the task provided a higher fidelity bounding box around the training
instance and thus a higher quality result. Table \ref{tab:quality-quantity}
shows that while the answers can begin with many errors, rollover
helps correct them and that the mean squared error of the final answer
is significantly better than the one-shot labeling.

\begin{table}[H]
\begin{tabular}{|c|c|c|}
\hline 
 & One-Shot Labeling & With Rollover (Final)\tabularnewline
\hline 
\hline 
MSE from baseline & 518.6px & 341.5px\tabularnewline
\hline 
Standard Deviation & 22.73px & 18.4px\tabularnewline
\hline 
\end{tabular}\\

\caption{Quality-Speed Tradeoff between one-shot labeling and labeling with
rollover}
\label{tab:quality-quantity}
\end{table}

\section{Discussion \& Future Work}

The conclusions of this experiment show that to unify a cognitive
architecture that incorporates interactive perception-action loops
and interactions between the robot and crowds of people on the Internet,
there is a clear need to reason about the speed at which you expect
a response (if the robot is presently interacting with a scene) and
the expectation of what quality of a response you can expect. Even
in the best circumstance, one-shot labeling can take at least 12.74
seconds and take upwards of 4 minutes depending on the crowd response
latency. Timing and scheduling when and to which interaction partners
the query must be made is an interesting dynamic challenge for this
always-on hybrid architecture. 

In performing this experiment, other strategies have made themselves
salient. For example, we found that while some of the bounding boxes
and labels that were provided are rife with error, on average, Amazon
Mechanical Turk workers provided good results. Our assumption was
that one-shot labeling will not on average provide good results and
that rollover would be needed, but instead, we found that with many
answers, a researcher can statistically remove outliers based on the
results of many parallel queries. This opens up the possibility to
an alternate strategy from strict rollover and to parallel one-shot
queries that may allow the agent to rely on the wisdom of many to
remove the untrustworthiness of smaller individuals, and then to simultaneously
model these workers and weight them appropriately. This parallel process
may simply capitalize on one-shot labeling speed and still keep high
quality results.

As we build social, embodied agents, we stake out a position that
emphasizes that many of the ``usable representations'' must have
social utility. These robots will inevitably be left to their own
devices at some point in their life cycle and while we can rely on
the idea that unsupervised learning can save us, this work emphasizes
the idea that robots are never alone when they are connected to the
Internet. It is a simple trade-off to begin to have interactions with
participants on-line. Additionally, we believe that scaling our research
from individual participants to groups of people to crowds of individuals
allows us to begin to investigate social reasoning at a larger scale,
opening up the idea of relationships between particular robotic agents
in the field and particular individuals online, and opening up the
possibility that an autonomous agent can begin to model the characteristics
of individuals and reason about their reliability at a larger scale
than ever thought possible with simple human-robot interaction alone.

\section*{Acknowledgements}

This work was partially funded by the MIT Media Lab Consortia, and
by NSF Grant CCF-1138986. We want to also thank Hae Won Park for her
many comments on earlier drafts of this paper, Jeff Orkin for his
creativity and Rob Miller for his leadership in pushing a crowdsourcing
agenda that has been inspiring to everyone involved with the Cambridge
Crowdfoo Initiative. 

\bibliographystyle{ieeetr}
\bibliography{../attention}

\end{document}